%% file: G2S_ICIP.tex
\documentclass{article}

\usepackage{spconf}
\usepackage{amsmath}
\usepackage{graphicx}
\usepackage{enumerate}
\usepackage{color}
\usepackage{soul}
\usepackage{amsfonts}
\usepackage{amssymb}
\usepackage{caption}
\usepackage{subcaption}
\usepackage{float}
\usepackage{comment}
\usepackage{url}

\DeclareMathOperator*{\argmin}{arg\,min}
\renewcommand{\vec}{\mathbf{vec}}

\title{ROBUST SURFACE RECONSTRUCTION FROM GRADIENTS VIA ADAPTIVE DICTIONARY REGULARIZATION}
\name{Andrew J. Wagenmaker, Brian E. Moore, and Raj Rao Nadakuditi
\thanks{This work was supported in part by the following grants: ONR grant N00014-15-1-2141, DARPA Young Faculty Award D14AP00086, and ARO MURI grants W911NF-11-1-0391 and 2015-05174-05.}}
\address{Department of EECS, University of Michigan, Ann Arbor, MI, USA}

\begin{document}

\maketitle

\begin{abstract}
This paper introduces a novel approach to robust surface reconstruction from photometric stereo normal vector maps that is particularly well-suited for reconstructing surfaces from noisy gradients. Specifically, we propose an adaptive dictionary learning based approach that attempts to simultaneously integrate the gradient fields while sparsely representing the spatial patches of the reconstructed surface in an adaptive dictionary domain. We show that our formulation learns the underlying structure of the surface, effectively acting as an adaptive regularizer that enforces a smoothness constraint on the reconstructed surface. Our method is general and may be coupled with many existing approaches in the literature to improve the integrity of the reconstructed surfaces. We demonstrate the performance of our method on synthetic data as well as real photometric stereo data and evaluate its robustness to noise.
\end{abstract}


\section{Introduction} \label{sec:intro}
\input{sec_intro.tex}

\section{Surface Reconstruction from Gradient Fields} \label{sec:background}
\input{sec_background.tex}

\section{Adaptive Dictionary Learning Regularization} \label{sec:dl}
\input{sec_dl.tex}

\section{Results} \label{sec:results}
\input{sec_results.tex}

\section{Conclusion}
In this work, we explored the use of adaptive dictionary learning based regularization for the estimation of surfaces from their gradient fields. We showed that our proposed dictionary learning approach is able to effectively reject the addition of noise to gradient fields/images and produce more accurate and smooth representations of the underlying surfaces compared to existing methods. Our dictionary learning framework is very general and would be straightforward to combine with many existing algorithms. In future work, we plan to investigate these combinations and perform a more thorough study of the influence of the various parameters of our dictionary learning model on the reconstructed surfaces.

\bibliographystyle{IEEEbib}
\ninept
\bibliography{G2S_ICIP}

\end{document}

%% file: sec_intro.tex

Imaging techniques such as photometric stereo \cite{woodham1980} allow one to efficiently estimate the normal vector map of an object. The primary goal of such methods is to ultimately derive a three-dimensional representation of the object, which requires some flavor of numerical integration of the gradient fields defined by the normal vector map. Robust photometric stereo---the problem of accurately determining the normal map of a non-ideal surface or from noisy data---has attracted considerable attention in recent years \cite{wu2011,ikehata2012,ikehata2014}. In this work, we seek to develop a robust approach to the problem of reconstructing surfaces from gradient fields that can accurately estimate a 3D representation of an object in the presence of noise.

The problem of reconstructing a surface from estimates of its photometric stereo gradient fields has been investigated since the late 1980s. The seminal works of Simchony \textit{et al.} \cite{simchony1990} and Frankot and Chellappa \cite{frankot1988} seek to solve the problem through a least squares approach, utilizing efficient discrete Fourier transform (DFT) or discrete cosine transform (DCT) based solvers.
Harker and O'Leary \cite{harker2008} propose a modified ``global'' least squares problem and extend this method to incorporate regularization \cite{harker2015}, solving a Sylvester equation to obtain the solution. Recently, Qu\'{e}au and Durou \cite{queau2015} introduced a weighted-least squares formulation as well as formulations minimizing total-variation and incorporating the $\ell_1$ norm to promote sparsity. Further attempts at applying a regularization term while integrating the gradients have also been proposed at the expense of computation time \cite{agrawal2006,ng2010}. Additional approaches include line-integral based methods \cite{wu1988,robles2005} and reconstructions based on the calculus of variations \cite{horn1986,balzer2012,durou2007}. A range of other methods have also been proposed with mixed results \cite{horovitz2004,lee1993,karaccali2003,karaccali2004,kovesi2005,balzer2011}.

Our work builds on these previous works, specifically those that utilize a least squares-type formulation to relate the underlying surface and its gradient fields. In particular, we propose a novel adaptive dictionary learning framework that enables the robust estimation of surfaces from noisy gradients. Dictionary learning \cite{elad2006image,aharon2006rm,kreutz2003dictionary} has, in recent years, been successfully applied to many imaging applications, e.g.,  \cite{ravishankar2011mr,ravishankar2016lassi,ravishankar2016low}. In dictionary learning models, one typically seeks to learn sparse representations of local regions (patches) of the data. These models often induce a type of smoothness constraint on the underlying data that, in the case of surface reconstruction, we show leads to robust reconstructions with desirable noise rejection properties. Our framework is general and can be easily combined with any existing method that utilizes a least squares-type objective to estimate the underlying surface.

%% file: sec_background.tex

Let $n(x,y) \in \mathbb{R}^3$ denote the normal vector of a differentiable surface $z(x,y)$ at position $(x,y)$, and let $n_1(x,y)$, $n_2(x,y)$, and $n_3(x,y)$ denote the $x$, $y$, and $z$ components of this vector, respectively. Under this ideal model, one can relate the $x$ and $y$ derivatives of the surface $z$ to its normal vectors via the relation
\begin{equation} \label{eq:partials}
\frac{\partial z(x,y)}{\partial x} = -p(x,y), \ \ \ \frac{\partial z(x,y)}{\partial y} = q(x,y),
\end{equation}
where we have defined $p(x,y) := n_1(x,y)/n_3(x,y)$ and $q(x,y) := n_2(x,y)/n_3(x,y)$. In practice, the estimated (e.g., by photometric stereo) normal vectors of a surface and its gradient fields will not exactly satisfy \eqref{eq:partials}, so one must instead find a function $z(x,y)$ with derivatives \emph{close} to $p(x,y)$ and $q(x,y)$ in an appropriate sense, often by minimizing a variational problem of the form
\begin{equation} \label{eq:cont_cost}
\int \int_{\Omega} \left (\frac{\partial z(x,y)}{\partial x}  - p(x,y) \right )^2 + \left ( \frac{\partial z(x,y)}{\partial y} - q(x,y) \right )^2 \ dx \ dy.
\end{equation}
When our data is instead sampled on a discrete grid, we will not have access to a continuous normal map $n(x,y)$ but will instead have a matrix $N \in \mathbb{R}^{m \times n \times 3}$ containing the normal vectors of the object on the grid. Following \eqref{eq:partials}, we can compute matrices $P \in \mathbb{R}^{m \times n}$ and $Q \in \mathbb{R}^{m \times n}$ containing the measured gradients, and our goal then becomes to estimate the matrix $Z \in \mathbb{R}^{m \times n}$ containing the values of the surface $z(x,y)$ sampled on the grid. The discrete analogue of \eqref{eq:cont_cost} is commonly expressed \cite{simchony1990,frankot1988,harker2008} as a standard least squares problem of the form
\begin{equation} \label{eq:surf_ls}
z^* = \argmin_{z} \ \|A z - v \|_2^2,
\end{equation}
where $z = \vec(Z) \in \mathbb{R}^{mn}$ is the vectorized surface, $A$ is a numerical differentiation operator, and the vector $v$ is an appropriate function of the measured gradients, $P$ and $Q$. Solving this problem yields a representation of our surface that is optimal in the least squares sense.

Note that the specific forms of $A$ and $v$ can vary. One possible formulation is
\begin{equation}
A = \begin{bmatrix}
D_n \otimes I_m \\
I_n \otimes D_m
\end{bmatrix}, \ \ \ \ v = \begin{bmatrix}
\textbf{vec}(P) \\
\textbf{vec}(Q)
\end{bmatrix},
\end{equation}
where $D_n$ is the discrete first differences matrix, and $\otimes$ denotes the Kronecker product. However, multiple models are possible, each yielding reconstructed surfaces with different properties. Importantly, the dictionary learning based approach that we introduce in Section~\ref{sec:dl} can be coupled with any least squares model of the form \eqref{eq:surf_ls}, so our proposed approach is quite flexible.

%% file: sec_dl.tex

Given normal vectors corrupted by noise or other non-idealities, solving \eqref{eq:surf_ls} directly generally produces a rough, bumpy surface, even when the underlying true surface is smooth. Thus, in this work, we propose an adaptive dictionary regularizer that can be combined with the least squares model \eqref{eq:surf_ls} to more accurately estimate the underlying surface. In particular, we propose to solve the dictionary learning problem
\begin{align} \label{eq:dl_surf}
\argmin_{z,B,D} & ~ \frac{1}{2} \left \| A z - v \right \|_2^2 + \lambda \bigg(\sum_{j = 1}^c \left \| P_j z - D b_j \right \|_2^2 + \mu^2 \left \| B \right \|_0 \bigg) \nonumber \\
\text{s.t.} & ~ \left \| d_i \right \|_2 = 1, \ \ \left \| b_j \right \|_{\infty} \leq a, \ \ \forall i,j.
\end{align}
Here, $P_j$ is a patch extraction operator that extracts a vectorized $c_x \times c_y$ spatial patch from $z$, $D \in \mathbb{R}^{c_x c_y \times K}$ is a dictionary matrix whose columns $d_i$ are vectorized $c_x \times c_y$ atoms, and $B \in \mathbb{R}^{K \times c}$ is a matrix of sparse codes, where column $b_j$ of $B$ defines the (usually sparse) linear combination of atoms used to represent the patch $P_j z$ of $z$. Also, $\left \| \ . \ \right \|_0$ is the familiar $\ell_0$ ``norm" and $\lambda, \mu > 0$ are regularization parameters. 

We include the constraints $\|b_j\|_{\infty} \leq a$, where $a$ is typically very large, to prevent pathologies that could theoretically arise during minimization since \eqref{eq:dl_surf} is non-coercive with respect to $B$, but the constraint is inactive in practice \cite{sairajfes2}. In addition, we constrain the columns of $D$ to be unit-norm to avoid scaling ambiguity between $D$ and $B$ \cite{kar}. Note that \eqref{eq:dl_surf} is a non-convex problem.

By solving \eqref{eq:dl_surf}, we are attempting to estimate our surface $z$ by numerically integrating its gradient fields through a least squares functional while simultaneously enforcing that local patches of the reconstructed surface should have sparse representations with respect to the dictionary $D$. As $D$ itself is learned, our proposed algorithm can automatically adapt to the underlying properties of the surface and its gradients. Intuitively, since the same dictionary of atoms is used to (sparsely) represent all patches, the atoms can learn universal features of the surface. This effectively equips \eqref{eq:dl_surf} with global information to estimate each patch, which (as we will show) can yield robust reconstructions when the gradients are noisy/corrupted.

\begin{figure*}
\centering
\begin{subfigure}[b]{0.18\textwidth}
  \includegraphics[width=\textwidth]{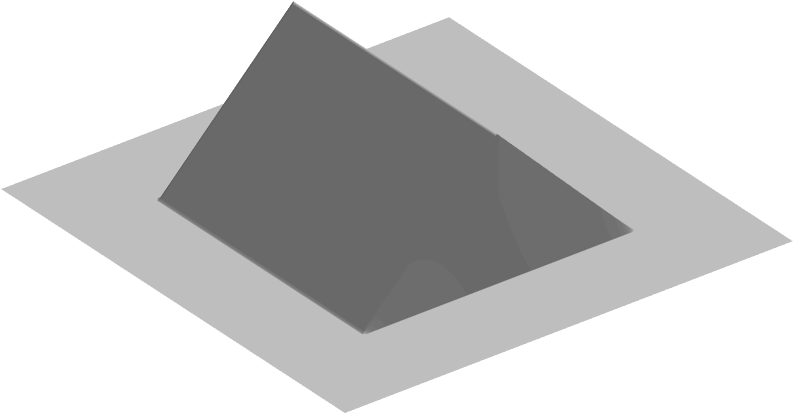}
  \caption{Ground Truth}
\end{subfigure}
~ 
\begin{subfigure}[b]{0.18\textwidth}
  \includegraphics[width=\textwidth]{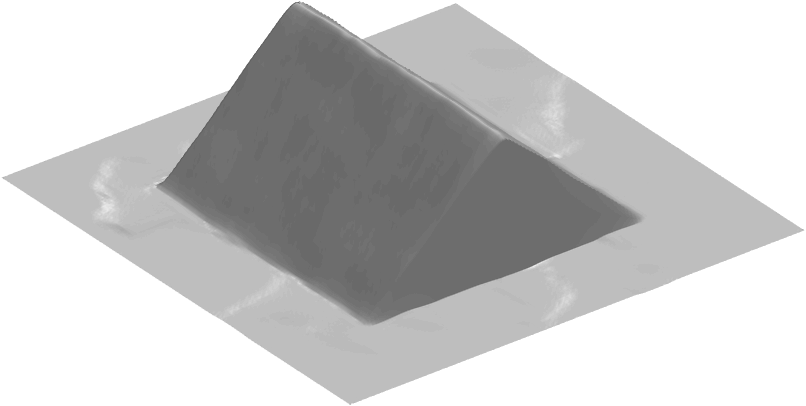}
  \caption{\textbf{DLS}}
\end{subfigure}
~ 
\begin{subfigure}[b]{0.18\textwidth}
  \includegraphics[width=\textwidth]{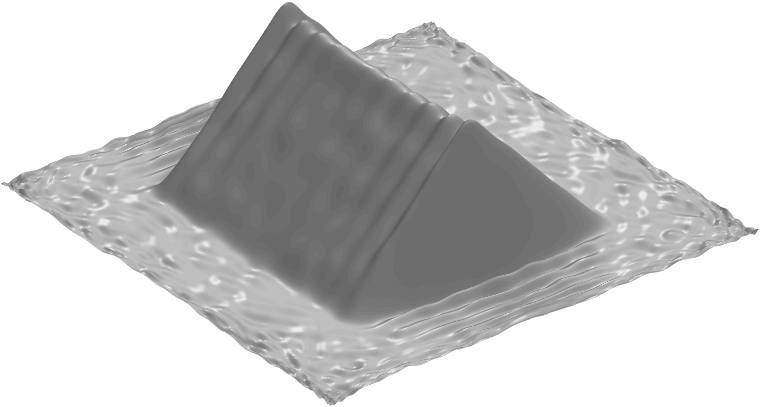}
  \caption{SR}
\end{subfigure}
~ 
\begin{subfigure}[b]{0.18\textwidth}
  \includegraphics[width=\textwidth]{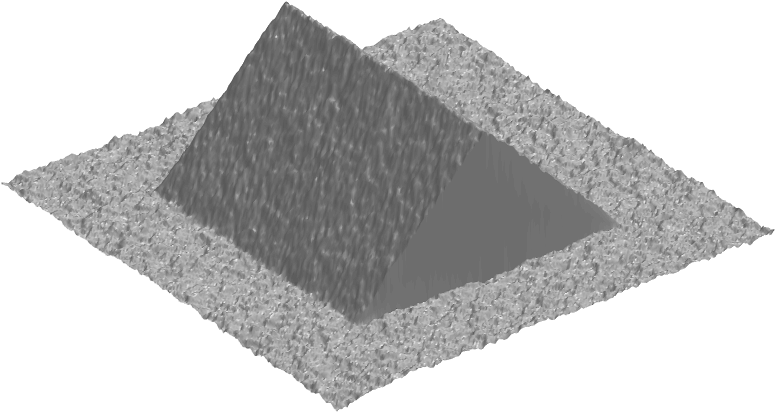}
  \caption{TV}
\end{subfigure}
~
\begin{subfigure}[b]{0.18\textwidth}
  \includegraphics[width=\textwidth]{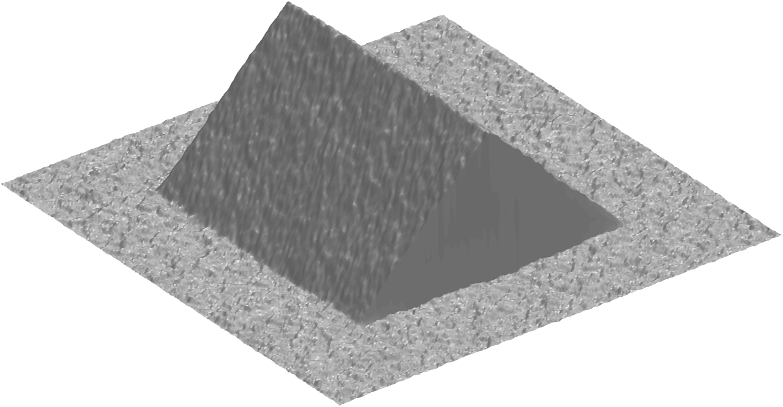}
  \caption{DCTLS}
\end{subfigure}
\caption{Surface reconstructions of the Tent dataset with SNR = 20 dB.}
\label{fig:tent}
\end{figure*}

\begin{figure*}
\centering
\begin{subfigure}[b]{0.18\textwidth}
  \includegraphics[width=\textwidth]{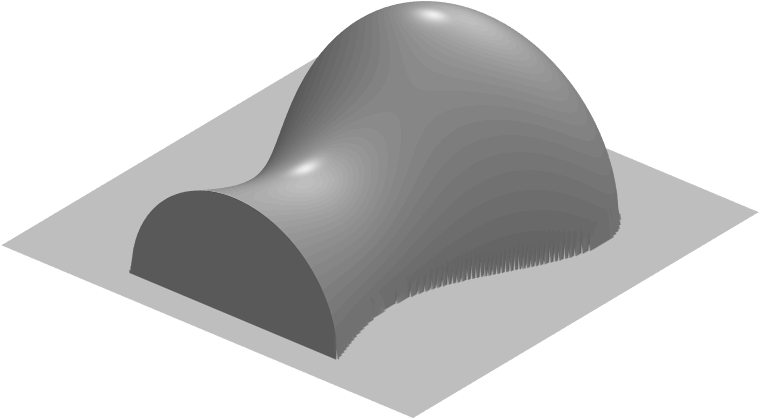}
  \caption{Ground Truth}
\end{subfigure}
~ 
\begin{subfigure}[b]{0.18\textwidth}
  \includegraphics[width=\textwidth]{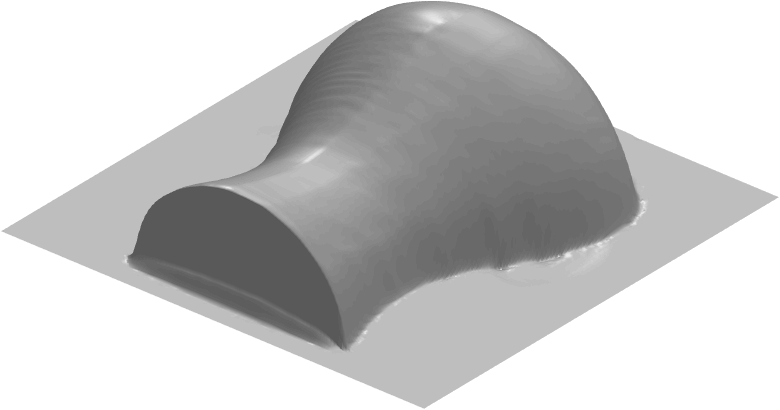}
  \caption{\textbf{DLS}}
\end{subfigure}
~ 
\begin{subfigure}[b]{0.18\textwidth}
  \includegraphics[width=\textwidth]{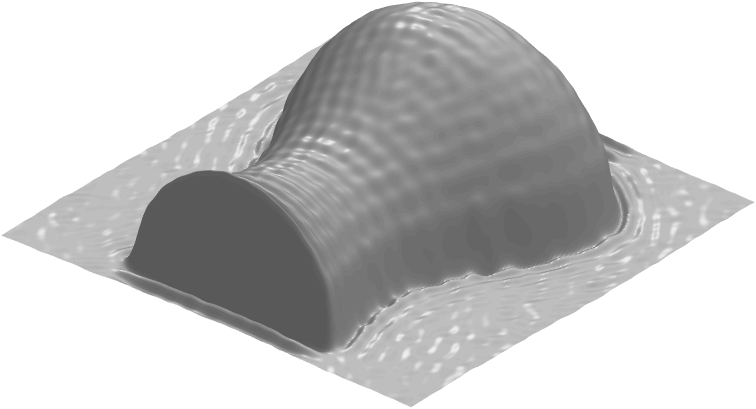}
  \caption{SR}
\end{subfigure}
~ 
\begin{subfigure}[b]{0.18\textwidth}
  \includegraphics[width=\textwidth]{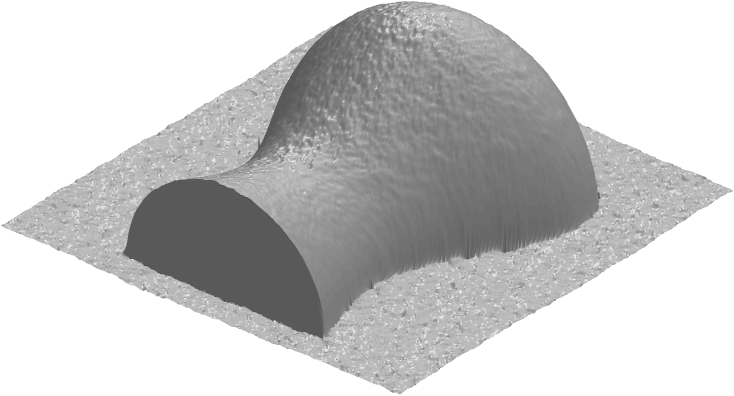}
  \caption{TV}
\end{subfigure}
~
\begin{subfigure}[b]{0.18\textwidth}
  \includegraphics[width=\textwidth]{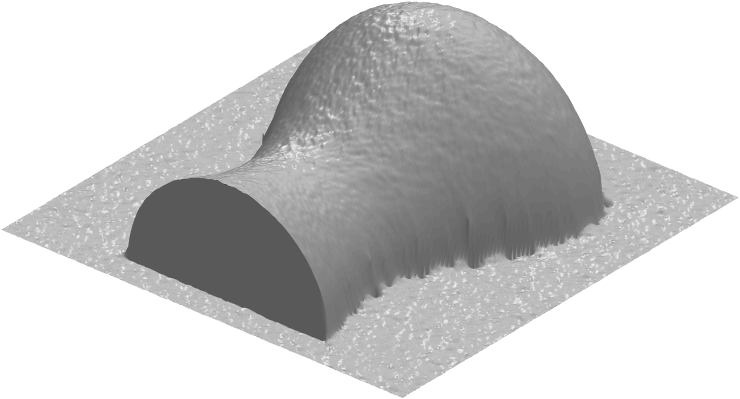}
  \caption{DCTLS}
\end{subfigure}
\caption{Surface reconstructions of the Vase dataset with SNR = 30 dB.}
\label{fig:vase}
\end{figure*}

\subsection{Dictionary Learning on Surfaces (DLS) Algorithm}
We propose to solve \eqref{eq:dl_surf} via a block coordinate descent-type algorithm where we alternate between updating $z$ with $(D,B)$ fixed and updating $(D,B)$ with $z$ fixed. Henceforward, we refer to this algorithm as the Dictionary Learning on Surfaces (DLS) method. 

\subsubsection{(D, B) updates}

Let $P$ be the matrix with columns $P_j z$. With $z$ fixed, the minimization of \eqref{eq:dl_surf} with respect to $(D,B)$ is
\begin{equation}  \label{dbupdate}
\begin{array}{rl}
\displaystyle\min_{B,D} & \left \| P - DB \right \|_F^2 +  \mu^2 \left \| B \right \|_0 \vspace{0.1cm} \\
\text{s.t.} & \left \| d_i \right \|_2 = 1, \ \ \left \| b_j \right \|_{\infty} \leq a, \ \forall i,j.
\end{array}
\end{equation}
We solve \eqref{dbupdate} via a block coordinate descent method where we iteratively minimize the cost with respect to the $i$th row of $B$ and the $i$th atom, $d_i$, of $D$ for every $1 \leq i \leq K$ with all other variables held fixed. A full derivation of this step can be found in \cite{sairajfes2,dinokat2016}, which we omit here due to space considerations.

\subsubsection{$z$ update}
With $D$ and $B$ fixed, our problem becomes
\begin{equation} \label{nupdate}
\min_{z} \ \frac{1}{2} \left \| A z - v \right \|_2^2 \\ +  \lambda \sum_{j = 1}^c \left \| P_j z - D b_j \right \|_2^2.
\end{equation}
The cost in \eqref{nupdate} can be written in the form $f(z)+g(z)$, where 
\begin{equation}
f(z) = \frac{1}{2} \left \| A z - v \right \|_2^2, \ \ \ g(z) = \lambda \sum_{j = 1}^c \left \| P_j z - D b_j \right \|_2^2.
\end{equation}
We utilize a proximal gradient \cite{parboyd} strategy to solve \eqref{nupdate}, iteratively updating $z$ according to
\begin{equation} \label{nproxupdate}
z^{k+1} = \textbf{prox}_{\tau g}(z^{k} - \tau \nabla f(z^{k})),
\end{equation}
where
\begin{equation} \label{nprox}
\textbf{prox}_{\tau g} (y) := \argmin_{x} \ \frac{1}{2} \left \| y - x \right \|_2^2 +  \tau g(x).
\end{equation}
Defining $\tilde{z}^{k} := z^{k} - \tau \nabla f(z^{k})$, we see that \eqref{nproxupdate} and \eqref{nprox} imply that $z^{k+1}$ satisfies the normal equation
\begin{equation}\label{znormal}
\bigg(I + 2 \tau \lambda \sum_{j=1}^c P_j^T P_j \bigg) z^{k+1} = \tilde{z}^{k} + 2 \tau \lambda \sum_{j=1}^c P_j^T D b_j.
\end{equation}
The matrix on the left hand side of \eqref{znormal} is diagonal, so its inverse can be cheaply computed to solve for $z^{k+1}$. Note that \eqref{nupdate} is a simple least squares problem and, as such, could be minimized by other methods (e.g., conjugate gradients).

%% file: sec_results.tex

\begin{figure*}
\centering
\begin{subfigure}[b]{0.23\textwidth}
  \includegraphics[width=\textwidth]{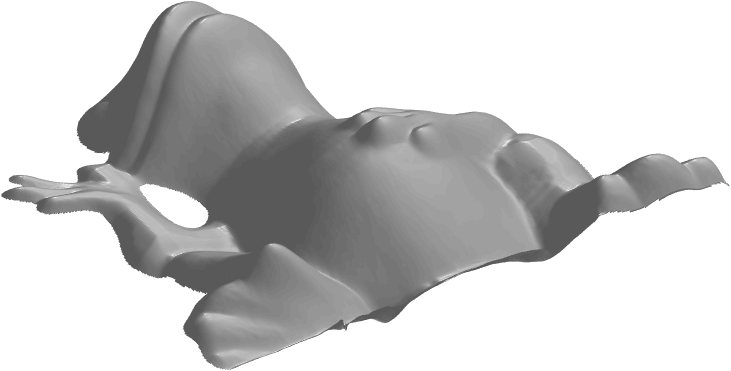}
\end{subfigure}
~ 
\begin{subfigure}[b]{0.23\textwidth}
  \includegraphics[width=\textwidth]{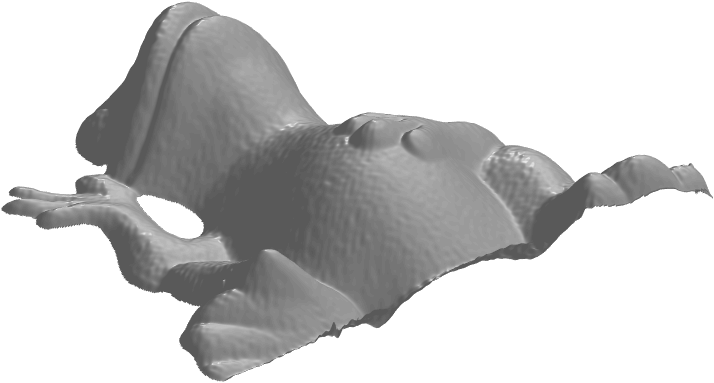}
\end{subfigure}
~ 
\begin{subfigure}[b]{0.23\textwidth}
  \includegraphics[width=\textwidth]{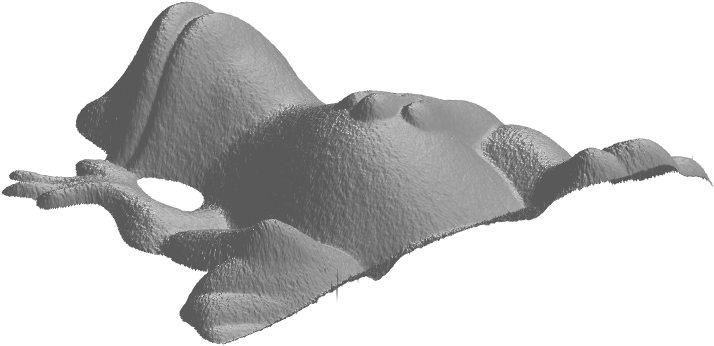}
\end{subfigure}
~
\begin{subfigure}[b]{0.23\textwidth}
  \includegraphics[width=\textwidth]{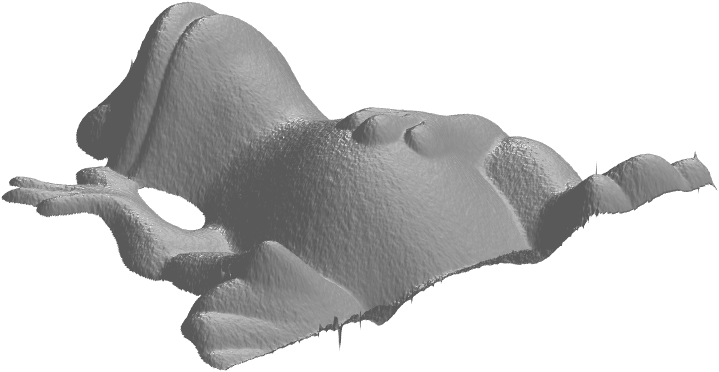}
\end{subfigure}
\end{figure*}

\setcounter{figure}{2}  

\begin{figure*}
\centering
\begin{subfigure}[b]{0.23\textwidth}
  \includegraphics[width=\textwidth]{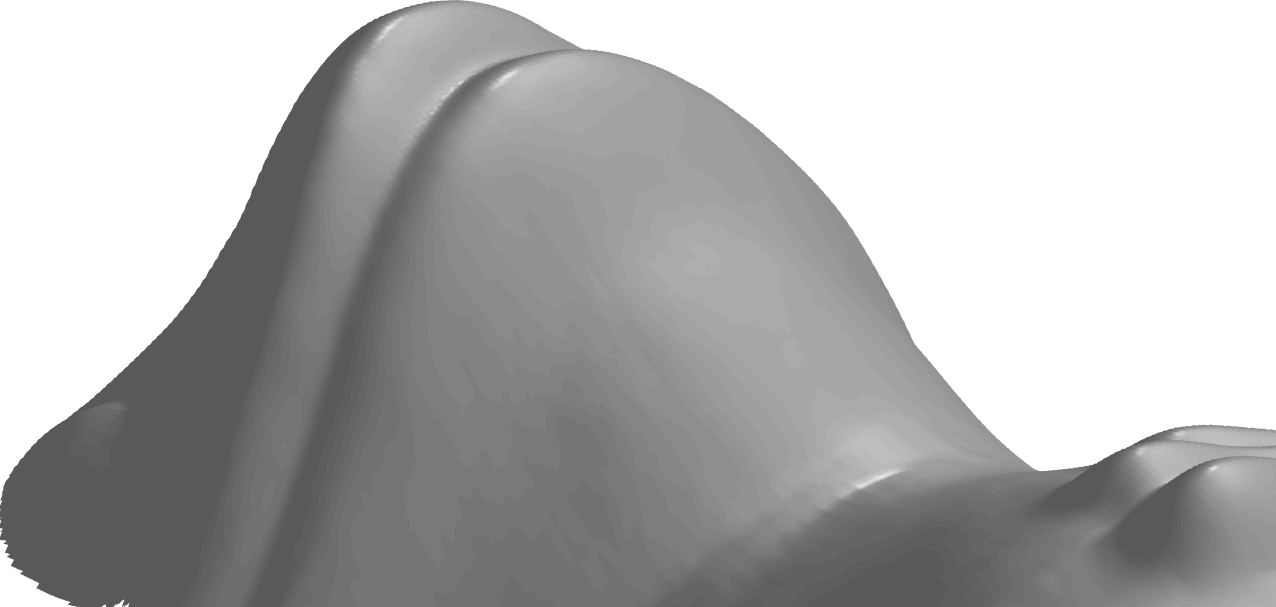}
  \caption{\textbf{DLS}}
\end{subfigure}
~ 
\begin{subfigure}[b]{0.23\textwidth}
  \includegraphics[width=\textwidth]{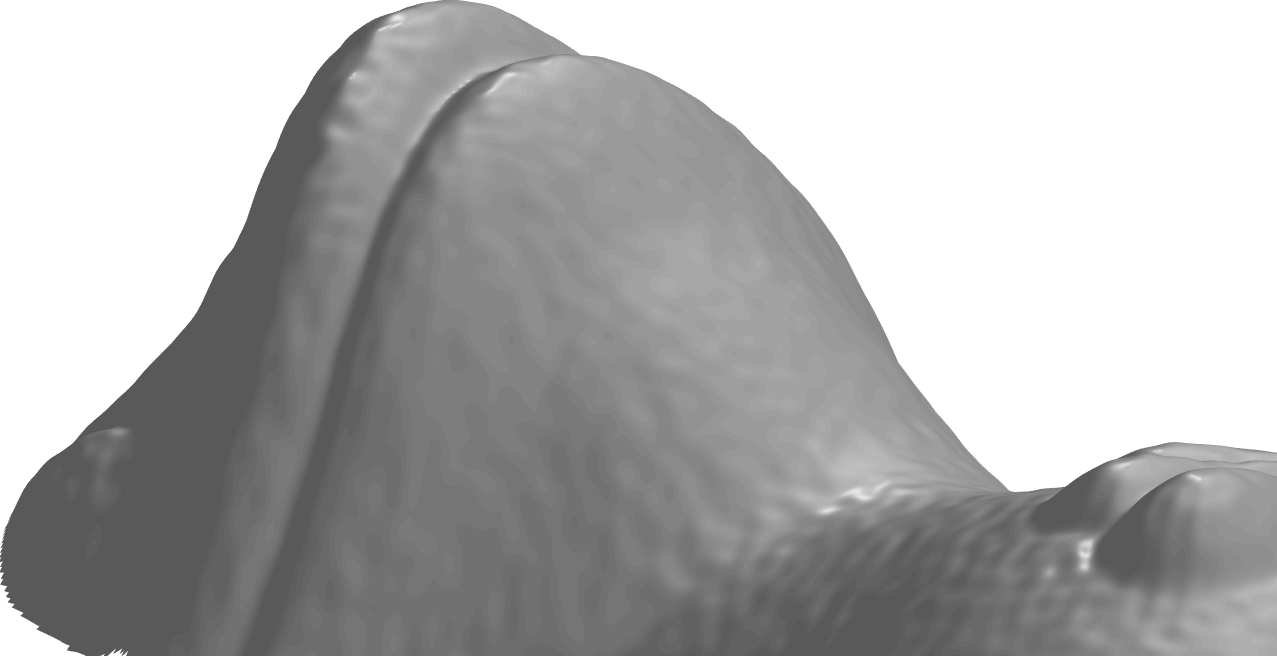}
  \caption{SR}
\end{subfigure}
~ 
\begin{subfigure}[b]{0.23\textwidth}
  \includegraphics[width=\textwidth]{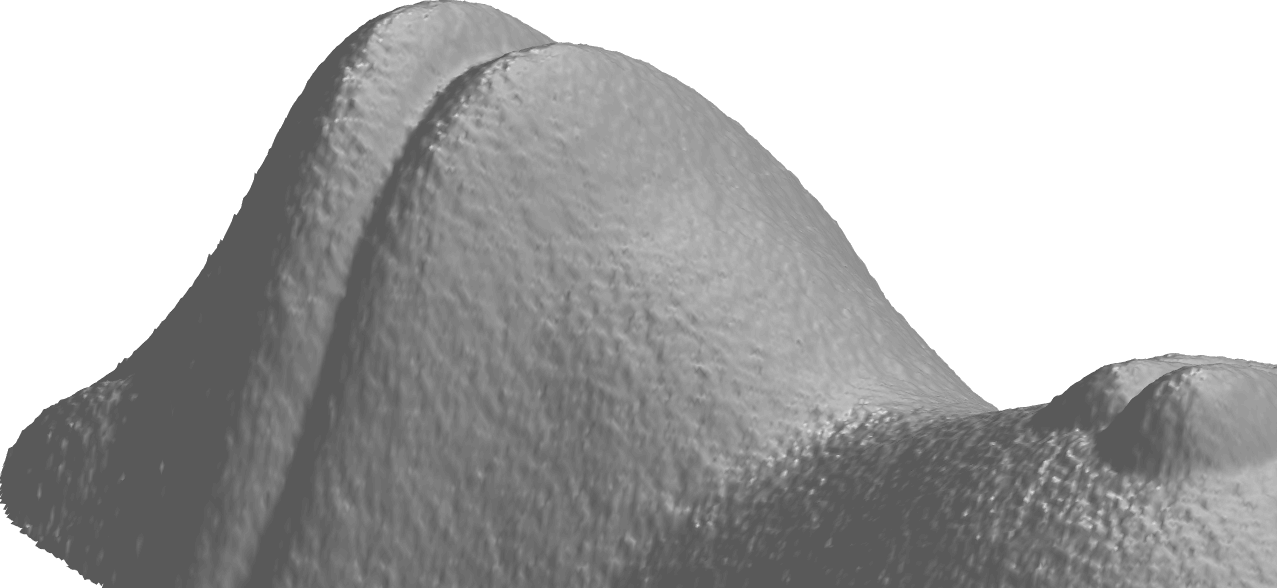}
  \caption{TV}
\end{subfigure}
~
\begin{subfigure}[b]{0.23\textwidth}
  \includegraphics[width=\textwidth]{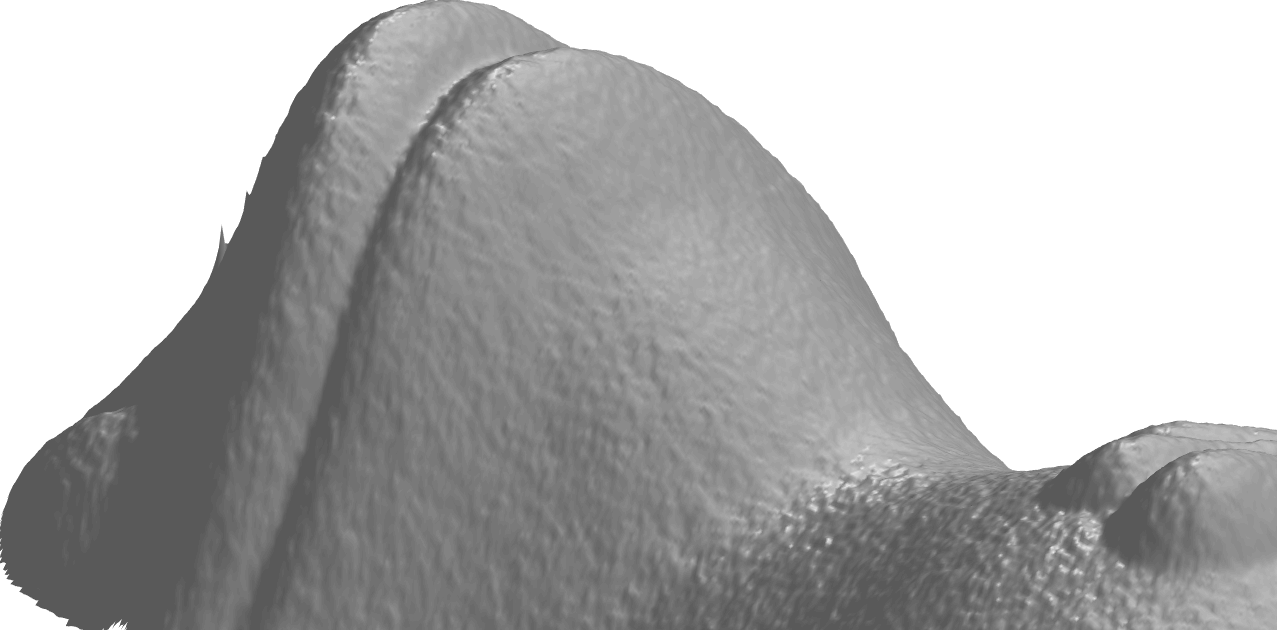}
  \caption{DCTLS}
\end{subfigure}
\caption{Surface reconstructions of the photometric stereo Frog dataset \cite{xiong2015shading} with SNR = 17 dB.}
\label{fig:frog_zoomed}
\end{figure*}

In this section, we numerically evaluate our proposed DLS method on several datasets. In each case we compare our method to the spectral regularization method (SR) \cite{harker2015}, the isotropic total variation (TV) method \cite{queau2015}, and DCT based least squares (DCTLS) \cite{simchony1990}. For methods that include tunable parameters, we sweep over a wide range of values, reporting the best results obtained. Our proposed DLS method can incorporate any least squares based solver by simply defining $A$ and $v$ in \eqref{eq:dl_surf} accordingly. For all results given here, we use the least squares cost found in \cite{simchony1990}. To evaluate the robustness of each algorithm, we add Gaussian noise to the data.

For our proposed DLS method, we use dictionary atoms of size $8 \times 8$ pixels and a square $64 \times 64$ dictionary $D$, initialized with a DCT matrix. We extracted patches from $z$ using a spatial stride of two pixels in each direction, allowing adjacent patches to overlap. Finally, we initialized $z$ as the vectorized surface produced by solving the stand-alone least squares problem in \cite{simchony1990}, and initialized $B = 0$.

\subsection{Synthetic Surface Reconstructions}
To quantitatively evaluate our method, we first considered two synthetic datasets, which we call ``Tent'' and ``Vase'', for which we have analytic expressions for $z = f(x,y)$. Given $f(x,y)$, we can differentiate to obtain the gradients, $\partial f(x,y) / \partial x$ and $\partial f(x,y) / \partial y$, and sample on a discrete grid. After reconstructing the surface from these gradients subject to additive noise, we evaluate the integrity of the reconstructions against the ground truth, $f(x,y)$, via the SSIM metric \cite{wang2004}. In these experiments, we add noise directly to the gradient fields to achieve a prescribed signal-to-noise ratio (SNR).

Figures~\ref{fig:tent} and \ref{fig:vase} show the reconstructed surfaces produced by each algorithm. As these images illustrate, the proposed DLS method produces significantly smoother surfaces from noisy data compared to the existing methods. Intuitively, the locally sparse model imposed by the dictionary regularization denoises the surfaces, while the adaptive nature of the dictionary allows DLS to represent and reconstruct both sharp edges and smooth regions on a data-dependent basis.

The surfaces obtained by SR, TV, and DCTLS are much more sensitive to the noisy gradients. Indeed, while they retain the general shape of the surface, they exhibit significantly more corruption. In particular, the spectral regularization method seems to introduce a systematic ``rippling" into the reconstructions.
\vspace{2mm}

\begin{table}[ht!]
\centering
\begin{tabular}{|c|c|c|c|c|}
\cline{1-5}
SNR (dB) & \bf{DLS} & SR & TV & DCTLS \\
\cline{1-5}
1 & \textbf{0.969} & 0.944 & 0.918 & 0.924 \\ 
\cline{1-5}
5 & \textbf{0.971} & 0.950 & 0.938 & 0.944 \\ 
\cline{1-5}
10 & \textbf{0.976} & 0.956 & 0.957 & 0.962 \\ 
\cline{1-5}
20 & \textbf{0.988} & 0.969 & 0.979 & 0.983 \\ 
\cline{1-5}
30 & \textbf{0.995} & 0.978 & 0.989 & 0.992 \\ 
\cline{1-5}
40 & \textbf{0.997} & 0.985 & 0.994 & 0.996 \\ 
\cline{1-5}
50 & \textbf{0.998} & 0.988 & 0.996 & \textbf{0.998} \\ 
\cline{1-5}
60 & \textbf{0.999} & 0.989 & 0.997 & 0.998 \\ 
\cline{1-5}
\end{tabular}
\caption{Quality of Tent reconstructions in SSIM as a function of SNR.}
\label{tab:pyramid}
\end{table}

\begin{table}[ht!]
\vspace{-4mm}
\centering
\begin{tabular}{|c|c|c|c|c|}
\cline{1-5}
SNR (dB) & \bf{DLS} & SR & TV & DCTLS \\
\cline{1-5}
1 & \textbf{0.958} & 0.930 & 0.889 & 0.894 \\ 
\cline{1-5}
5 & \textbf{0.966} & 0.934 & 0.911 & 0.915 \\ 
\cline{1-5}
10 & \textbf{0.971} & 0.942 & 0.933 & 0.936 \\ 
\cline{1-5}
20 & \textbf{0.977} & 0.961 & 0.965 & 0.966 \\
\cline{1-5} 
30 & \textbf{0.982} & 0.975 & 0.981 & 0.981 \\ 
\cline{1-5}
40 & \textbf{0.990} & 0.982 & 0.990 & 0.989 \\ 
\cline{1-5}
50 & \textbf{0.993} & 0.984 & \textbf{0.993} & 0.992 \\ 
\cline{1-5}
60 & \textbf{0.995} & 0.985 & \textbf{0.995} & 0.993 \\ 
\cline{1-5}
\end{tabular}
\caption{Quality of Vase reconstructions in SSIM as a function of SNR.}
\label{tab:vase}
\end{table}

\vspace{-2mm}
Tables~\ref{tab:pyramid} and \ref{tab:vase} numerically corroborate the qualitative results from Figures~\ref{fig:tent} and \ref{fig:vase}. In the low SNR regime, DLS significantly outperforms the other approaches. As SNR increases, the gap decreases. When the data is essentially noiseless, DLS, TV, and DCTLS are all able to reconstruct the surfaces with comparably negligible error.

\subsection{Photometric Stereo}
We now return to the problem of reconstructing a 3D representation of an object from normal vectors obtained through photometric stereo. We consider a dataset containing 10 images, each taken under a unique, known lighting direction, and corrupt the images with Gaussian noise at a prescribed SNR. We compute the normal vectors from the noisy images using the standard least squares approach \cite{wu2011}. Given the normal vectors, we then compute gradient fields as discussed in Section 2, and from those we generate a 3D reconstruction of the object using each method. Figure~\ref{fig:frog_zoomed} illusrates the results of this procedure on the Frog dataset,\footnote{This dataset can be found at \url{http://vision.seas.harvard.edu/qsfs/Data.html}} which contains real images of a frog statue \cite{xiong2015shading}.

The reconstructions in Figure~\ref{fig:frog_zoomed} showcase the ability of the proposed DLS approach to produce a smoother surface from the noisy gradients compared to the existing methods. The denoising capability of DLS may prove valuable when running photometric stereo on real-world data, where noise and other non-idealities are inevitable.

%% file: G2S_ICIP.bbl
\begin{thebibliography}{10}

\bibitem{woodham1980}
R.~J. Woodham,
\newblock ``Photometric method for determining surface orientation from
  multiple images,''
\newblock {\em Optical Engineering}, vol. 19, no. 1, pp. 191139--191139, 1980.

\bibitem{wu2011}
L.~Wu, A.~Ganesh, B.~Shi, Y.~Matsushita, Y.~Wang, and Y.~Ma,
\newblock ``Robust photometric stereo via low-rank matrix completion and
  recovery,''
\newblock in {\em ACCV}, 2010, pp. 703--717.

\bibitem{ikehata2012}
S.~Ikehata, D.~Wipf, Y.~Matsushita, and K.~Aizawa,
\newblock ``Robust photometric stereo using sparse regression,''
\newblock in {\em CVPR}, 2012, pp. 318--325.

\bibitem{ikehata2014}
S.~Ikehata and K.~Aizawa,
\newblock ``Photometric stereo using constrained bivariate regression for
  general isotropic surfaces,''
\newblock in {\em CVPR}, June 2014, pp. 2187--2194.

\bibitem{simchony1990}
T.~Simchony, R.~Chellappa, and M.~Shao,
\newblock ``Direct analytical methods for solving poisson equations in computer
  vision problems,''
\newblock {\em IEEE PAMI}, vol. 12, no. 5, pp. 435--446, May 1990.

\bibitem{frankot1988}
R.~T. Frankot and R.~Chellappa,
\newblock ``A method for enforcing integrability in shape from shading
  algorithms,''
\newblock {\em IEEE PAMI}, vol. 10, no. 4, pp. 439--451, 1988.

\bibitem{harker2008}
M.~Harker and P.~O'Leary,
\newblock ``Least squares surface reconstruction from measured gradient
  fields,''
\newblock in {\em CVPR}, 2008, pp. 1--7.

\bibitem{harker2015}
M.~Harker and P.~O'Leary,
\newblock ``Regularized reconstruction of a surface from its measured gradient
  field,''
\newblock {\em Journal of Mathematical Imaging and Vision}, vol. 51, no. 1, pp.
  46--70, 2015.

\bibitem{queau2015}
Y.~Qu{\'e}au and J-D Durou,
\newblock ``Edge-preserving integration of a normal field: Weighted
  least-squares, {TV} and $l^1$ approaches,''
\newblock in {\em International Conference on Scale Space and Variational
  Methods in Computer Vision}, 2015, pp. 576--588.

\bibitem{agrawal2006}
A.~Agrawal, R.~Raskar, and R.~Chellappa,
\newblock ``What is the range of surface reconstructions from a gradient
  field?,''
\newblock in {\em ECCV}, 2006, pp. 578--591.

\bibitem{ng2010}
H-S Ng, T-P Wu, and C-K Tang,
\newblock ``Surface-from-gradients without discrete integrability enforcement:
  A gaussian kernel approach,''
\newblock {\em IEEE PAMI}, vol. 32, no. 11, pp. 2085--2099, 2010.

\bibitem{wu1988}
Z.~Wu and L.~Li,
\newblock ``A line-integration based method for depth recovery from surface
  normals,''
\newblock {\em Computer Vision, Graphics, and Image Processing}, vol. 43, no.
  1, pp. 53--66, 1988.

\bibitem{robles2005}
A.~Robles-Kelly and E.~R. Hancock,
\newblock ``A graph-spectral method for surface height recovery,''
\newblock {\em Pattern Recognition}, vol. 38, no. 8, pp. 1167--1186, 2005.

\bibitem{horn1986}
B.~K.~P. Horn and M.~J. Brooks,
\newblock ``The variational approach to shape from shading,''
\newblock {\em Computer Vision, Graphics, and Image Processing}, vol. 33, no.
  2, pp. 174--208, 1986.

\bibitem{balzer2012}
J.~Balzer and T.~M{\"o}rwald,
\newblock ``Isogeometric finite-elements methods and variational reconstruction
  tasks in vision---a perfect match,''
\newblock in {\em CVPR}, 2012, pp. 1624--1631.

\bibitem{durou2007}
J-D Durou and F.~Courteille,
\newblock ``Integration of a normal field without boundary condition,''
\newblock in {\em PACV}, 2007, pp. 8--p.

\bibitem{horovitz2004}
I.~Horovitz and N.~Kiryati,
\newblock ``Depth from gradient fields and control points: Bias correction in
  photometric stereo,''
\newblock {\em Image and Vision Computing}, vol. 22, no. 9, pp. 681--694, 2004.

\bibitem{lee1993}
K.~M. Lee and C-C~J. Kuo,
\newblock ``Surface reconstruction from photometric stereo images,''
\newblock {\em JOSA A}, vol. 10, no. 5, pp. 855--868, 1993.

\bibitem{karaccali2003}
B.~Kara{\c{c}}al{\i} and W.~Snyder,
\newblock ``Reconstructing discontinuous surfaces from a given gradient field
  using partial integrability,''
\newblock {\em Computer Vision and Image Understanding}, vol. 92, no. 1, pp.
  78--111, 2003.

\bibitem{karaccali2004}
B.~Kara{\c{c}}al{\i} and W.~Snyder,
\newblock ``Noise reduction in surface reconstruction from a given gradient
  field,''
\newblock {\em International Journal of Computer Vision}, vol. 60, no. 1, pp.
  25--44, 2004.

\bibitem{kovesi2005}
P.~Kovesi,
\newblock ``Shapelets correlated with surface normals produce surfaces,''
\newblock in {\em ICCV}, 2005, vol.~2, pp. 994--1001.

\bibitem{balzer2011}
J.~Balzer,
\newblock ``A gauss-newton method for the integration of spatial normal fields
  in shape space,''
\newblock {\em Journal of Mathematical Imaging and Vision}, vol. 44, no. 1, pp.
  65--79, 2012.

\bibitem{elad2006image}
M.~Elad and M.~Aharon,
\newblock ``Image denoising via sparse and redundant representations over
  learned dictionaries,''
\newblock {\em IEEE Trans. on Image Proc.}, vol. 15, no. 12, pp. 3736--3745,
  2006.

\bibitem{aharon2006rm}
M.~Aharon, M.~Elad, and A.~Bruckstein,
\newblock ``$k$-svd: An algorithm for designing overcomplete dictionaries for
  sparse representation,''
\newblock {\em IEEE Trans. on Signal Proc.}, vol. 54, no. 11, pp. 4311--4322,
  2006.

\bibitem{kreutz2003dictionary}
K.~Kreutz-Delgado, J.~F. Murray, B.~D. Rao, K.~Engan, T-W Lee, and T.~J.
  Sejnowski,
\newblock ``Dictionary learning algorithms for sparse representation,''
\newblock {\em Neural Computation}, vol. 15, no. 2, pp. 349--396, 2003.

\bibitem{ravishankar2011mr}
S.~Ravishankar and Y.~Bresler,
\newblock ``{MR} image reconstruction from highly undersampled k-space data by
  dictionary learning,''
\newblock {\em IEEE Trans. on Med. Imag.}, vol. 30, no. 5, pp. 1028--1041,
  2011.

\bibitem{ravishankar2016lassi}
S.~Ravishankar, B.~E. Moore, R.~R. Nadakuditi, and J.~A. Fessler,
\newblock ``Lassi: A low-rank and adaptive sparse signal model for highly
  accelerated dynamic imaging,''
\newblock in {\em IVMSP Workshop}, 2016, pp. 1--5.

\bibitem{ravishankar2016low}
S.~Ravishankar, B.~Moore, R.~R. Nadakuditi, and J.~Fessler,
\newblock ``Low-rank and adaptive sparse signal {(LASSI)} models for highly
  accelerated dynamic imaging,''
\newblock {\em IEEE Trans. on Med. Imag.}, 2017.

\bibitem{sairajfes2}
S.~Ravishankar, R.~R. Nadakuditi, and J.~A. Fessler,
\newblock ``Efficient sum of outer products dictionary learning {(SOUP-DIL)} -
  the $\ell_{0}$ method,''
\newblock {\em arXiv preprint arXiv:1511.08842}, 2015.

\bibitem{kar}
R.~Gribonval and K.~Schnass,
\newblock ``Dictionary identification--sparse matrix-factorization via
  $\textit{l}_{1}$-minimization,''
\newblock {\em IEEE Trans. on Inform. Theory}, vol. 56, no. 7, pp. 3523--3539,
  2010.

\bibitem{dinokat2016}
S.~Ravishankar, B.~E. Moore, R.~R. Nadakuditi, and J.~A. Fessler,
\newblock ``Efficient learning of dictionaries with low-rank atoms,''
\newblock in {\em Proc. IEEE Global Conference on Signal and Information
  Processing}, 2016, pp. 222--226.

\bibitem{parboyd}
N.~Parikh and S.~Boyd,
\newblock ``Proximal algorithms,''
\newblock {\em Found. Trends Optim.}, vol. 1, no. 3, pp. 127--239, Jan. 2014.

\bibitem{xiong2015shading}
Y.~Xiong, A.~Chakrabarti, R.~Basri, S.~J. Gortler, D.~W. Jacobs, and T.~E.
  Zickler,
\newblock ``From shading to local shape.,''
\newblock {\em IEEE PAMI}, vol. 37, no. 1, pp. 67--79, 2015.

\bibitem{wang2004}
Z.~Wang, A.~C. Bovik, H.~R. Sheikh, and E.~P. Simoncelli,
\newblock ``Image quality assessment: from error visibility to structural
  similarity,''
\newblock {\em IEEE Trans. on Image Proc.}, vol. 13, no. 4, pp. 600--612, 2004.

\end{thebibliography}
